\documentclass[letterpaper]{article} 
\usepackage{aaai24}  
\usepackage{times}  
\usepackage{helvet}  
\usepackage{courier}  
\usepackage[hyphens]{url}  
\usepackage{graphicx} 
\usepackage{natbib}  
\usepackage{caption} 
\usepackage{algorithm}
\usepackage{algorithmic}

\usepackage{color-edits}
\usepackage{booktabs}
\usepackage{multirow}
\usepackage{tabularx}
\usepackage{caption}
\usepackage{subcaption}
\usepackage{newfloat}
\usepackage{listings}
\DeclareCaptionStyle{ruled}{labelfont=normalfont,labelsep=colon,strut=off} 
\floatstyle{ruled}
\newfloat{listing}{tb}{lst}{}
\floatname{listing}{Listing}
\title{Where It Really Matters: Few-Shot Environmental Conservation Media Monitoring for Low-Resource Languages}
\author{
    Sameer Jain\textsuperscript{\rm 1},
    Sedrick Scott Keh\textsuperscript{\rm 1}, 
    Shova Chettri\textsuperscript{\rm 2},
    Karun Dewan\textsuperscript{\rm 2},\\
    Pablo Izquierdo\textsuperscript{\rm 2},
    Johanna Prussman\textsuperscript{\rm 2},
    Pooja Shreshtha\textsuperscript{\rm 2}, 
    César Suárez\textsuperscript{\rm 2},\\
    Zheyuan Ryan Shi\textsuperscript{\rm 3}, 
    Lei Li\textsuperscript{\rm 1}, 
    Fei Fang\textsuperscript{\rm 1}\\
}
\affiliations{
    \textsuperscript{\rm 1}Carnegie Mellon University \\
    \textsuperscript{\rm 2}World Wide Fund for Nature \\ 
    \textsuperscript{\rm 3}University of Pittsburgh
    \\

}

\usepackage{bibentry}

\begin{document}

\maketitle

\begin{abstract}
Environmental conservation organizations routinely monitor news content on conservation in protected areas to maintain situational awareness of developments that can have an environmental impact. Existing automated media monitoring systems require large amounts of data labeled by domain experts, which is only feasible at scale for high-resource languages like English. However, such tools are most needed in the global south where news of interest is mainly in local low-resource languages, and far fewer experts are available to annotate datasets sustainably. In this paper, we propose NewsSerow, a method to automatically recognize environmental conservation content in low-resource languages. NewsSerow is a pipeline of summarization, in-context few-shot classification, and self-reflection using large language models (LLMs).
Using at most 10 demonstration example news articles in Nepali, NewsSerow significantly outperforms other few-shot methods and achieves comparable performance with models fully fine-tuned using thousands of examples. 
The World Wide Fund for Nature (WWF) has deployed NewsSerow for media monitoring in Nepal, significantly reducing their operational burden, and ensuring that AI tools for conservation actually reach the communities that need them the most. NewsSerow has also been deployed for countries with other languages like Colombia.
\end{abstract}

\section{Introduction}
News monitoring for environmental conservation has long been a significant task of many nonprofit organizations. The organizations aim to remain informed about developments that can have potentially significant environmental impacts, and intervene promptly, such as advocating for more environmental-friendly progress.
However, manually reading and analyzing news articles is a very labor-intensive and time-consuming process. Recent studies have eased this burden by automating the scraping, classification, and analysis of conservation-related articles~\cite{keh-2023-newspanda}. Such efforts allow for early detection of high-impact activities and enable nonprofits to take timely action. 

However, although effective, such approaches come with a hefty price tag: they require a lot of labeled data prepared by domain experts at these organizations, not only in training the model, but also in its maintenance after the tool has been deployed. This is significant extra work for the staff at these conservation organizations who are already occupied with their routine job. As such, it is no wonder previous works focus on articles written in English only, due to the relative abundant human resources. However, these news monitoring tools would actually make a much bigger difference in the global south where the news of interest is mainly in local low-resource languages.
The reason is two-fold: First, in these regions, there are more developments that may yield high environmental impacts. Second, many projects do not gain exposure on international media until they reach a later more developed stage, if they gain any exposure at all, where it would be too late for nonprofits to intervene. 
Our collaborators at the World Wide Fund for Nature (WWF), a conservation-focused nonprofit, have local offices in over 70 countries. 
Admittedly, for many of these 70+ countries, it is impossible to develop and maintain a media monitoring tool in local languages in the same way as has been done in English -- the operational burden and data work are too much for their limited workforce. 

From a technical perspective, automatically classifying news articles written in local languages is challenging because many of these are low-resource languages that do not enjoy the same power from large language models (LLMs) as they are trained on high-resource languages like English. Cultural factors may also come into play: certain things that are acceptable in Western countries may be taboo under certain cultures, and vice-versa. Consequently, as we will show later, off-the-shelf language models may struggle to accurately classify articles in low-resource languages. Fine-tuning is also difficult because it requires a neatly annotated dataset, which is often not readily available. 

To tackle this challenge, we propose NewsSerow, a multilingual few-shot classification method for conservation news monitoring. NewsSerow is built on top of LLMs and consists of a novel pipeline of summarization, in-context few-shot demonstrations, and self-reflection.
For news articles written in Nepali, with only 10 examples, NewsSerow significantly outperforms other few-shot methods and can even achieve comparable performance with models fully fine-tuned using more than 1800 labeled data points. NewsSerow was initially developed and demonstrated for Nepali, and hence the algorithm's name -- Himalayan serow, an important animal in Nepal. The method is scalable and language-agnostic, allowing it to generalize well to a wide variety of languages with relatively little effort. Beyond the results for Nepali, we also show positive results for Spanish. Our system is currently being deployed by the country offices of WWF in Nepal and Colombia.


\section{Related Work}
\paragraph{News Monitoring Pipelines} The earliest published work similar to NewsSerow was that of \citet{hosseini-coll-ardanuy}. They collected and labelled an English dataset, then used BERT models \cite{devlin-2019-bert} to classify articles as positive or negative to conservation. However, the work was not deployed in the real world. Meanwhile, \citet{keh-2023-newspanda} collect localized datasets for English articles covering India and Nepal. Their proposed toolkit focuses specifically on conservation and infrastructure projects and has been deployed in collaboration with WWF. However, none of these previous studies address the news articles written in local languages, which is a crucial gap that our system aims to fill.

\paragraph{Multilingual Models} Adapting large multilingual models such as m-BERT \cite{devlin-2019-bert} and XLM-R~\cite{conneau-2020-unsupervised} to specific downstream tasks through fine-tuning is a popular approach in multilingual settings. More recently, larger models in the scale of tens or hundreds of billions of parameters \cite{brown-2020-language, OpenAI2023GPT4TR, Touvron2023Llama2O} have shown remarkable ability to understand and generate coherent text in a plethora of languages, including low-resource languages such as Nepali. These massive models are notably able to operate in zero-shot or few-shot settings. As such, they become the foundation in the development of NewsSerow, for which we rely on few-shot in-context learning for domain-specific classification, utilizing domain expert insights along with recent techniques in prompting. 

\paragraph{Prompting}{Prompting has emerged as a major tool to elicit reasoning abilities from language models \cite{DBLP:journals/corr/abs-2107-13586}. This usually involves providing the model with an instruction or a set of few-shot examples, then having the model complete the input prompt. Chain-of-thought reasoning \cite{wei-2022-chain, Kojima2022LargeLM} involves asking language models to reason in a scratchpad-like manner \cite{scratchpad} before providing a final response. 

\paragraph{LLMs To Critique and Self-Correct} As the capabilities of language models continue to progress, strong models such as gpt-3.5-turbo or gpt-4 \cite{OpenAI2023GPT4TR} have been shown to be reasonable as evaluation tools \cite{alpaca_eval, Gilardi2023ChatGPTOC, jain-2023-multi} to evaluate other outputs as part of the inference pipeline. Further, they possess the ability to reflect and iterate over their own responses as well ~\cite{jang2023reflection, kadavath-2022-language}. By adding this layer of filtering, we can achieve nontrivial performance improvements \cite{shinn-2023-reflexion}, which is especially useful in structured generation applications such as code generation
\cite{Welleck2022GeneratingSB}. We incorporate a reflection step into the NewsSerow pipeline.

\section{NewsSerow Methodology}

\begin{figure*}
    \centering
    \includegraphics[width=\textwidth]{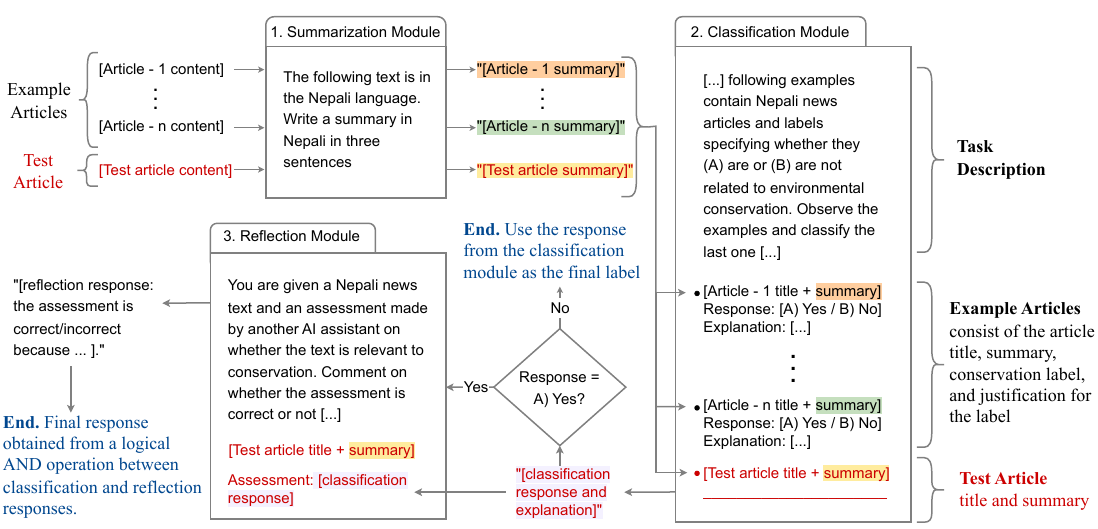}
    \caption{NewsSerow prompt pipeline. We illustrate the flow of the test example in red. Model responses that are used as input for later prompts are color-coded by background. For example, the test article summary (highlighted yellow) is generated by the summarization module and used in the classification and reflection modules.}
    \label{fig:main}
\end{figure*}

\subsection{Outline of the Pipeline}
Figure~\ref{fig:main} contains an overview of the NewsSerow pipeline.
Our framework uses a gpt-3.5-turbo model to perform classification. In our default setup, we utilize 10 randomly chosen demonstration examples for in-context learning. For each article in the test set, our pipeline consists of three steps. We list the steps before delving deeper in corresponding subsections:

\begin{enumerate}
    \item \textbf{Summarization:} We leverage the zero-shot generation capabilities of gpt-3.5-turbo to generate a summary of the test article in a given language.  
    \item \textbf{Classification:} We pass the model the titles and summaries of 10 randomly chosen example articles with their true labels along with the title and summary of the test article. We extract the preliminary conservation label from the model response at this stage.
    \item \textbf{Reflection:} We further prompt the model to determine the correctness of its own preliminary response by passing the test article again, but this time along with its own output from the previous step. 
\end{enumerate} 
This pipeline is general and works for other LLMs pretrained on multilingual data.

\subsection{Summarization Module} \label{subsec:summary}
We pass an article to gpt-3.5-turbo and direct it to generate a zero-shot summary in the same language. We prompt the model to restrict the summary to 3 sentences to prevent overflowing the context window in the next step.

Since our main classification step requires example articles to be fed into the prompt along with the test article, we cannot feed the entirety of articles due to the limit set by the gpt-3.5-turbo context window. Summarization allows us to efficiently utilize the limited context window of the model while ensuring that the salient points of the article are included in the prompt. We study the effect of this step in an ablation study in Section \ref{subsec:ablations}.

\subsection{Classification Module} \label{subsec:classification}
The classification module uses \textit{in-context learning} \cite{brown-2020-language}, which is the capability of LLMs to learn to perform a task by utilizing a few input-output examples.
In this step, we feed gpt-3.5-turbo a prompt that consists of:
\begin{enumerate}
    \item A task description that defines environmental conservation and instructs the LLM to observe the given examples and comment on whether the test example is relevant to conservation or not.
    \item A set of in-context examples (10, by default) that are randomly sampled from the training set. For each example article, we feed in the article title and summary, followed by a response specifying whether or not the article is relevant to conservation, along with an explanation for the classification.
    \item The title and summary of the test example.
\end{enumerate}
Following the pattern it observes in the examples, the model outputs a classification label for the test example along with a justification for the label.

In providing explanations for the true labels of the in-context examples, we utilize \textit{chain-of-thought} prompting \cite{wei-2022-chain}, which is the capacity of LLMs to observe intermediate reasoning steps in the given in-context examples to improve their own performance on reasoning tasks. In addition to improved performance, chain-of-thought reasoning gives the additional advantage of having the model justify its classification decision for the test example, which can be used by NGOs using the system as a source of additional distilled information. 

The classification module is followed by a decision step. If the article is classified to be not relevant to conservation, the pipeline ends here with \texttt{not relevant} being the final label. If the article is classified to be relevant, we go through an additional filtering step, described next.

\subsection{Reflection Module} \label{subsec:reflection}
False positives are one of the most frequent category of errors we observe when using in-context learning for classification. Furthermore, in the justifications given by the model for its decisions, the model often makes a tenuous connection between the article content and environmental conservation. Consequently, it labels the article as being relevant to conservation, thus driving up false positives. For example, the model often assigns a positive label to articles that mention protected areas, even in cases when the focus of the news is not conservation.

To address this, we add a \textit{reflection} module to our pipeline \cite{kadavath-2022-language, shinn-2023-reflexion}. Concretely, if the classification module labels an article as being relevant to conservation, we go through an additional filtering step. We pass the test example (title + summary) to a new prompt along with its own classification label from the previous step. We then ask the LLM to comment on the correctness of this assessment. If it thinks its previous positive assessment is correct, the pipeline would output \texttt{relevant}. Otherwise, \texttt{not relevant} would be the final label.


\section{Experiments}
\label{sec:experiments}
We run a series of experiments and ablation studies to demonstrate the effectiveness of NewsSerow. 
NewsSerow is a few-shot classification algorithm designed specifically for low-resource languages. As an example of such languages, we use its results on Nepali news articles as the primary performance indicator of NewsSerow. Of course, there is nothing that stops NewsSerow from working with other, high-resource languages. In fact, to demonstrate its generalizability, we also run experiments on Colombian news articles which are written in Spanish.

\subsection{Datasets} \label{subsec:data}
We collect a dataset of 796 news articles (548 Nepal and 248 Colombia) with binary labels for their relevance to environmental conservation. We collect the data for Colombia and Nepal using NewsAPI and a custom crawler respectively: 
\begin{enumerate}
\item \textbf{Colombian Data}: We use NewsAPI to obtain all Spanish language news articles. We obtain articles relevant to Colombia by filtering articles that either have the string ``Colombia" in the title or if they come from a Colombian news domain name. We increase the percentage of conservation-related news in the dataset by further filtering by a search term consisting of the name of a protected area in Colombia, taken from a list of relevant areas curated by WWF. We query NewsAPI for a 90-day period from December 2022 to March 2023 and obtain a set of 2748 articles, from which we sample 248 articles for labeling. The articles are labeled by domain experts from WWF for relevance to conservation.

\item \textbf{Nepali Data:} Since NewsAPI currently does not support the Nepali language, we built a custom web crawler to obtain Nepali news articles. We scrape articles from four Nepali news websites suggested by WWF and filter the articles by matching on a search term consisting of the name of a protected area in Nepal. Collection was done from December 2022 to March 2023. Articles were scraped weekly and labeled by domain experts from WWF for relevance to conservation. The final set contained 548 articles.
\end{enumerate}

In addition, we also have 1647 news articles in English with similar binary labels. These data were shared by the authors of~\cite{keh-2023-newspanda}. We use all of these English articles for training the Nepali NewsSerow and Colombian NewsSerow. Table~\ref{tab:dataset} describes the data used in the training, validation, and testing of NewsSerow.

\begin{table*}[t]
\centering
{
\begin{tabular}{ccc}
\toprule
\textbf{Split} & \textbf{Nepal/Nepali} & \textbf{Colombia/Spanish}\\
\midrule
Demonstrations (Few-shot) & 10 Nepali articles & 10 Spanish articles\\
Training (Fine-tuned) & 172 Nepali articles + 1647 English articles & 73 Spanish articles + 1647 English articles \\
Validation & 100 Nepali articles & 50 Spanish articles\\
Test & 276 Nepali articles & 125 Spanish articles\\

\bottomrule
\end{tabular}%
}
\caption{The train-validation-test split of the dataset used in the experiments shown in Table~\ref{tab:main}. For the English data in Translation Test (see Table~\ref{tab:main}), out of 1647 examples, we use 300 for test, 300 for validation, and the rest for training.}
\label{tab:dataset}
\end{table*}



\begin{table*}
\centering
{%
\begin{tabular}{lcccccc}
\toprule
\textbf{Model} &  \multicolumn{3}{c}{\textbf{Nepal/Nepali}} & \multicolumn{3}{c}{\textbf{Colombia/Spanish}} \\
\cmidrule(lr){2-4}
\cmidrule(lr){5-7}
& \textbf{Precision} & \textbf{Recall} & \textbf{F1-Score}  & \textbf{Precision} & \textbf{Recall} & \textbf{F1-Score}\\
\midrule
\textbf{Zero-shot Models} & & & & \\
\cmidrule{1-1}
GPT-3.5-Turbo & $0.79 \:(0.02)$
& $0.32 \:(0.01)$
& $0.46 \:(0.01)$
& $1.00 \:(0.00)$ 
& $0.14 \:(0.00)$ 
& $0.25 \:(0.00)$ \\
\midrule
\textbf{Few-shot Models} & & & & \\
\cmidrule{1-1}
mBERT & $0.26 \:(0.03)$ 
& $0.70 \:(0.12)$ 
& $0.37 \:(0.01)$ 
& $0.25 \:(0.03)$ 
& $0.71 \:(0.18)$ 
& $0.36 \:(0.04)$  \\

XLM-R & $0.35 \:(0.04)$ 
& $0.60 \:(0.23)$ 
& $0.43 \:(0.06)$  
& $0.49 \:(0.24)$ 
& $0.61 \:(0.41)$ 
& $0.44 \:(0.25)$ \\

GPT-3.5-Turbo & $0.68 \:(0.05)$ 
& $0.58 \:(0.07)$ 
& $0.62 \:(0.02)$ 
& $0.39 \:(0.03)$ 
& $1.00 \:(0.00)$ 
& $0.56 \:(0.03)$  \\

NewsSerow & $\underline{0.88} \:(0.03)$ 
& $\underline{0.58} \:(0.01)$ 
& $\underline{\mathbf{0.70}} \:(0.01)$  
& $\underline{0.89} \:(0.02)$ 
& $\underline{0.71} \:(0.00)$ 
& $\underline{\mathbf{0.79}} \:(0.01)$  \\
\midrule

\textbf{Fine-tuned Models} & & & & \\
\cmidrule{1-1}
mBERT & $0.77 \:(0.06)$ 
& $0.57 \:(0.05)$ 
& $0.65 \:(0.02)$  
& $0.86 \:(0.10)$ 
& $0.80 \:(0.17)$ 
& $\mathbf{0.81} \:(0.08)$  \\

XLM-R & $0.70 \:(0.05)$ 
& $0.71 \:(0.04)$ 
& $\mathbf{0.70}\: (0.01)$ 
& $0.69 \:(0.03)$ 
& $0.95 \:(0.03)$ 
& $0.80 \:(0.02)$ \\

Translation Test & $0.74 \:(0.16)$ 
& $0.64 \:(0.15)$ 
& $0.66 \:(0.06)$
& $0.76 \:(0.21)$ 
& $0.67 \:(0.19)$
& $0.66 \:(0.16)$ \\






\bottomrule
\end{tabular}%
}
\caption{We compare the performance of NewsSerow against a set of zero-shot, few-shot, and fully fine-tuned baselines. All metrics are averaged over 5 runs with standard deviations shown in parentheses. 
On F1-score, NewsSerow performs significantly better than the few-shot baselines for both languages. More importantly, it also matches the performance of the best fine-tuned model for Nepali while staying marginally below state-of-the-art performance for Colombian, while having much lower variance and using significantly less data.}
\label{tab:main}
\end{table*}

\subsection{Evaluation Metrics}
We model the identification of relevant news articles as a binary classification task and use precision, recall, and F1 score for evaluation. Since the dataset is imbalanced in favour of negative examples, we do not report accuracy. We report the metrics on the positive class. All the numbers we report are an average across 5 runs. For the mBERT and XLM-R baselines, we use 5 different random seeds for training. For the results on GPT models, we use 5 random seeds to shuffle in-context examples in the prompt. We set temperature to zero for all GPT experiments in order to make the responses deterministic.

\subsection{Baselines}
We compare the performance of NewsSerow to the following baselines:
\subsubsection{Zero-Shot Baselines}
\begin{enumerate}
    \item \textbf{Zero-Shot GPT-3.5-Turbo}: We prompt the off-the-shelf model to perform zero-shot classification, i.e., without using any in-context examples and also not using the other features of our proposed framework--chain-of-thought, zero-shot summarization, and reflection.
\end{enumerate}

\subsubsection{Few-Shot Baselines}
\begin{enumerate}
    \item \textbf{Few-Shot GPT-3.5-Turbo}: We prompt the model to use in-context learning with 10 examples but without using the other features of NewsSerow -- chain-of-thought, zero-shot summarization, and reflection. In place of the zero-shot summary, we use the first 3 sentences of the article. We discuss the effect of this choice in the ablation study (Section \ref{subsec:ablations}) on the summarization module.
    
    \item \textbf{Few-Shot Finetuned Multilingual Encoders}: We fine-tune large multilingual models such as mBERT \cite{devlin-2019-bert} and XLM-R \cite{conneau-2020-unsupervised}. We fine-tune a different model for each language (Nepali or Spanish) on 10 examples from that language. We use a batch size of 1, an initial learning rate equal to 5e-06 for the AdamW optimizer \cite{loshchilov-2018-decoupled}. We train for 30 epochs and use the best epoch to evaluate.
    
\end{enumerate}

\begin{figure}[t]
    \centering
    \begin{subfigure}[b]
    {\columnwidth}
        \centering
        \includegraphics[width=0.9\textwidth]{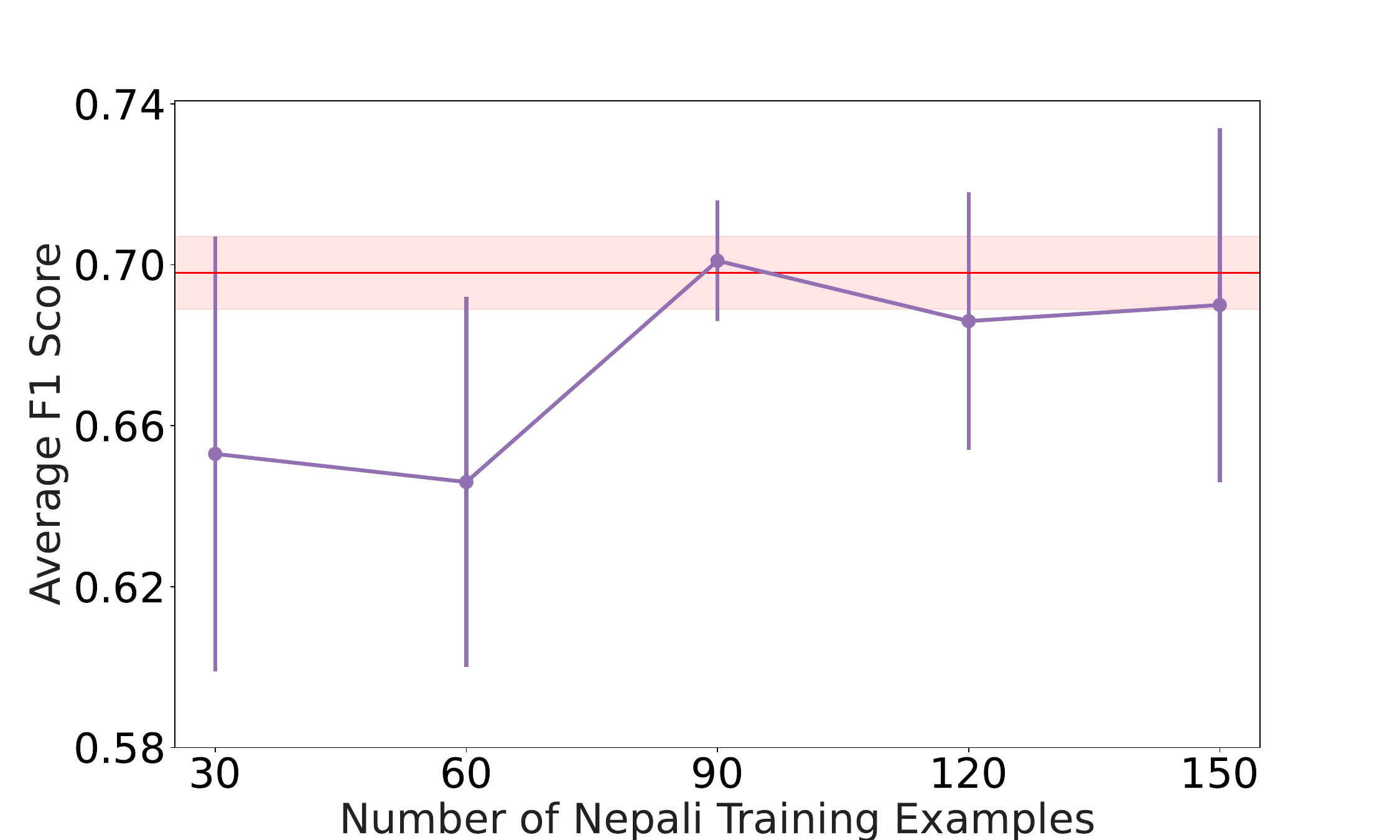}
        \caption{}
        \label{fig:number_ex_xlmr}
    \end{subfigure}

    \begin{subfigure}[b]{\columnwidth}
        \centering
        \includegraphics[width=0.9\textwidth]{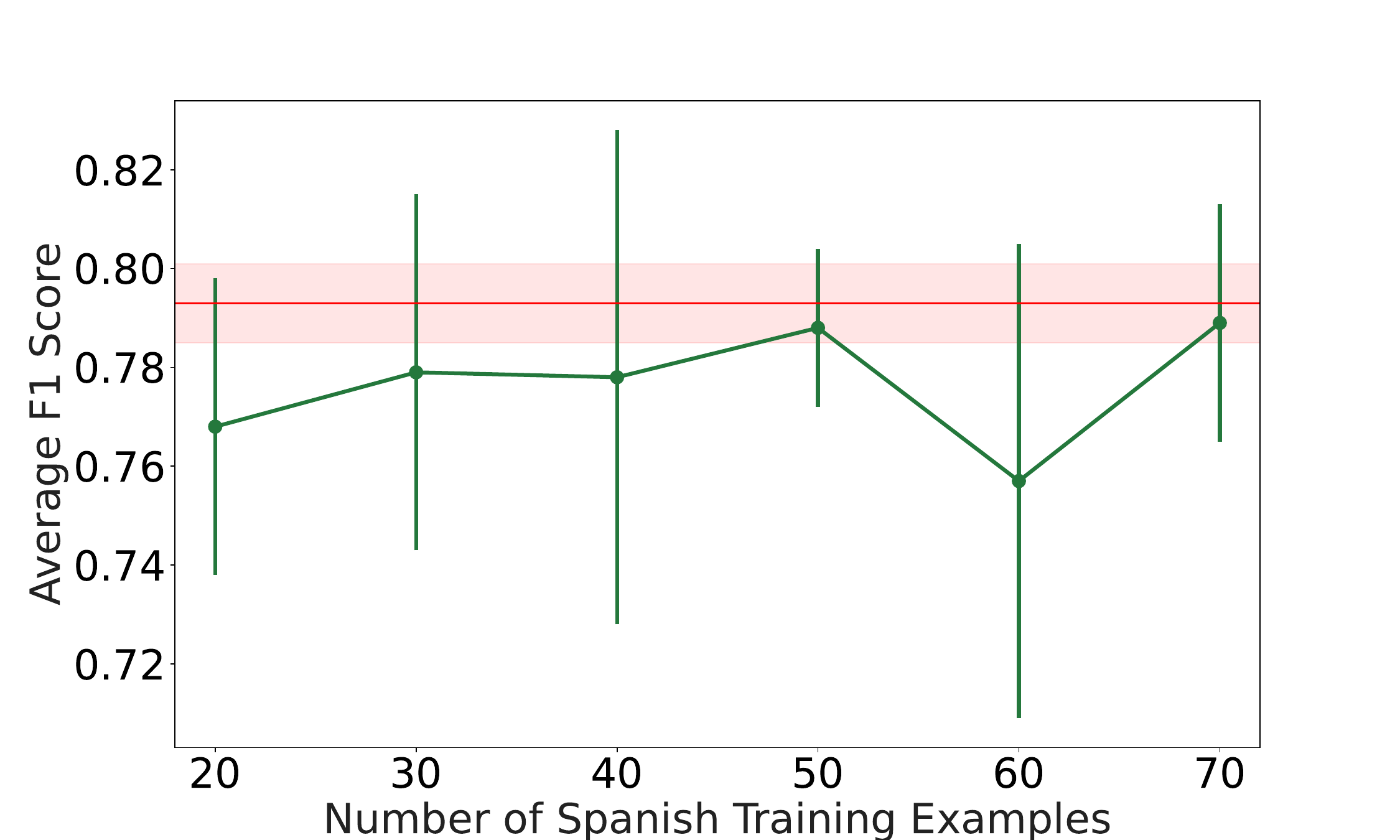}
        \caption{}
        \label{fig:num_examples_xlm_col.pdf}
    \end{subfigure}
    \caption{Fine-tuned XLM-R's performance with number of training examples on the (a) Nepali (purple) and (b) Spanish (green) test sets. X-axis shows the number of target language (Nepali/Spanish) training examples, in addition to which 1647 English examples are used to fine-tune the models. We show NewsSerow's 10-shot performance on the same test sets in red with a light red error bar.}
\end{figure}

\begin{figure}[t]
    \centering
    \includegraphics[width=240pt]{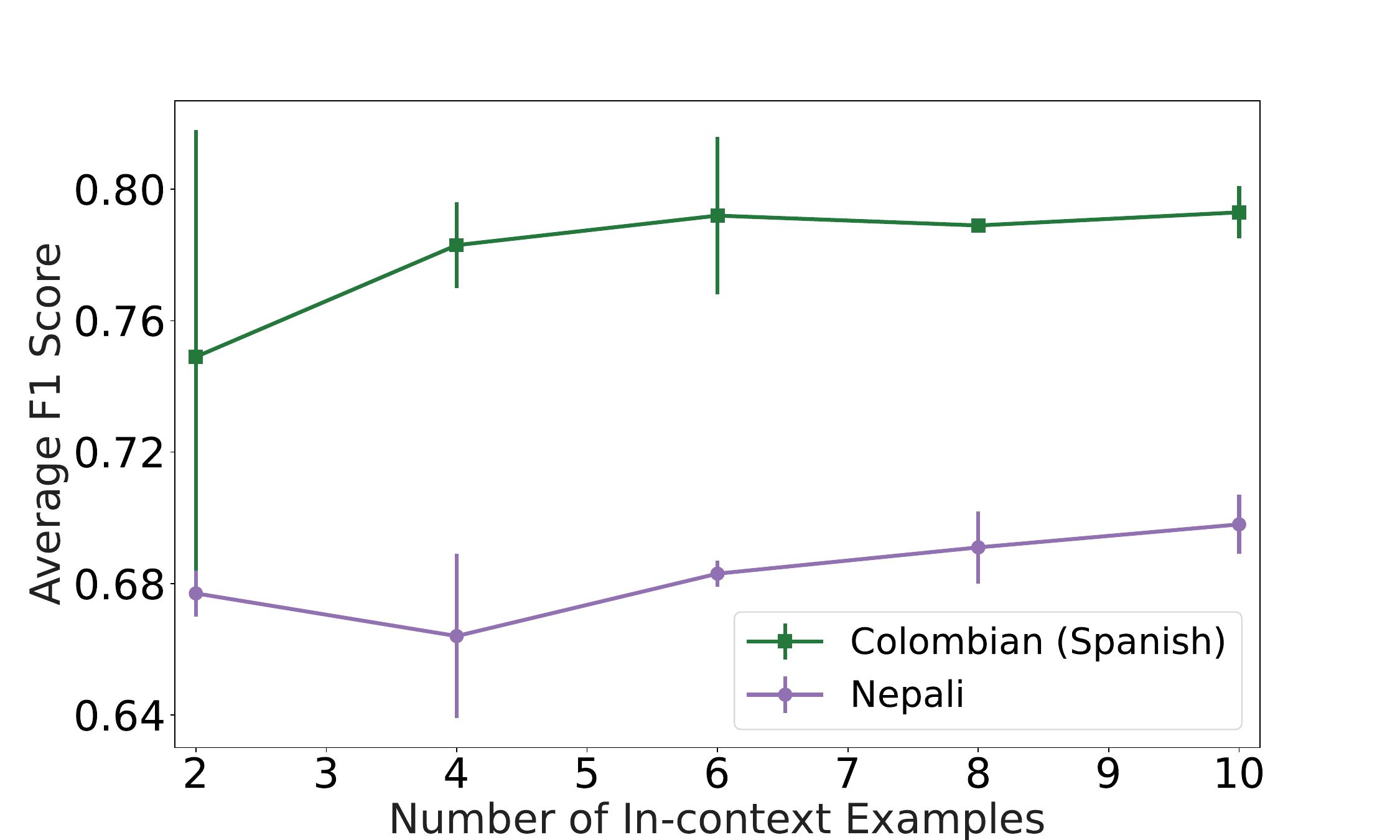}
    \caption{NewsSerow's performance with number of in-context examples}
    \label{fig:number_ice}
\end{figure}

\subsubsection{Fully Fine-Tuned Baselines}
\begin{enumerate}
    \item \textbf{Translate-Test with RoBERTa}: We have much more data~\cite{keh-2023-newspanda} in the English language as compared to the other languages. We fine-tune a RoBERTa \cite{liu-2019-roberta} model on the English data consisting of 1647 articles labeled for relevance to environmental conservation. We then use a translate-test pipeline to classify examples from the other languages.
    \item \textbf{Fully Fine-Tuned XLM-R and mBERT}: We fine-tune mBERT and XLM-R on all available training data. We have 172 Nepali training examples and 73 Spanish training examples. In addition to target language data, we find that using English language data boosts performance, so we include 1647 English articles in the training set, resulting in a total of 1819 training examples for Nepali and 1720 examples for Spanish. We use a batch size of 1, an initial learning rate equal to 5e-06 for the AdamW optimizer. We train for 10 epochs and use the best epoch to evaluate.

\end{enumerate}

\subsection{Results and Analyses}
\subsubsection{Comparing NewsSerow with Off-the-Shelf GPT-3.5-Turbo}
NewsSerow is a few-shot framework that uses, by default, 10 examples. We compare its performance with a vanilla GPT-3.5-Turbo model. We provide the vanilla model the same set of in-context examples without any of the prompt engineering techniques used in NewsSerow, such as the chain-of-thought explanations and the zero-shot summaries. We instead use the initial three sentences of the article as a summary \cite{narayan-2018-dont, nenkova-2005-automatic}.

As shown in Table~\ref{tab:main}, NewsSerow outperforms a vanilla GPT-3.5-Turbo model on both languages, gaining 8 points and 23 points on mean F1 Score on Nepali and Spanish respectively. It also reduces the variance compared to the vanilla model. For Nepali, NewsSerow improves precision by 20 points while not suffering a drop in recall.

\subsubsection{Comparing NewsSerow with Multilingual Encoders}
We compare NewsSerow with XLM-R and mBERT trained in two settings:

\begin{enumerate}
    \item \textbf{Few-Shot:}
    Since NewsSerow uses 10 examples, we first compare it to XLM-R and mBERT fine-tuned on the same 10 examples (Table~\ref{tab:main}). Both models vastly underperform NewsSerow in the few-shot setting. In fact, they also underperform a vanilla gpt-3.5-turbo model. This is unsurprising, given that LLMs like gpt-3.5-turbo are designed to generalize well from a few examples, while traditional models such as mBERT and XLM-R require more data-intensive finetuning.
    
    \item \textbf{Full Finetuning:} 
    Since XLM-R and mBERT are typically intended to be fine-tuned on larger datasets, we fine-tune them on all available training data, including English language data, which also boosts performance. In summary, we finetune on 1819 examples (172 Nepali + 1647 English) for the Nepali models and 1720 examples (73 Spanish + 1647 English) for the Spanish models. As illustrated in Table~\ref{tab:main}, NewsSerow matches or exceeds the performance of these models while using a fraction of the data. Furthermore, it does this while achieving strictly lower variance as compared to mBERT and XLM-R. For the translation-test results, while they outperform some of the few-shot baselines, they perform slightly worse than both the NewsSerow and the fine-tuned mBERT/XLM-R models. This is likely due to inaccuracies during the translation process. 
\end{enumerate}

\begin{table}[ht]
\small
\centering
{%
\begin{tabular}{cccccc}
\toprule
\multicolumn{3}{c}{\textbf{Setting}} &  \multicolumn{3}{c}{\textbf{Performance}} \\
\cmidrule(lr){1-3}
\cmidrule(lr){4-6}
\textbf{CoT.} & \textbf{Sum.} & \textbf{Ref.} & \textbf{Precision} & \textbf{Recall} & \textbf{F1 Score} \\
\midrule

No & No & No & 
$0.68 \:(0.05)$ & 
$0.58 \:(0.07)$ & 
$0.62 \:(0.02)$\\

No & No & Yes & 
$0.94 \:(0.02)$ & 
$0.28 \:(0.01)$ & 
$0.43 \:(0.01)$\\

No & Yes & No & 
$0.70 \:(0.04)$ & 
$0.68 \:(0.06)$& 
$0.69 \:(0.03)$\\

No & Yes & Yes & 
$0.91 \:(0.02)$ & 
$0.49 \:(0.01)$ & 
$0.63 \:(0.01)$\\

Yes & No & No & 
$0.57 \:(0.05)$ & 
$0.82 \:(0.05)$ &
$0.67 \:(0.02)$\\

Yes & No & Yes & 
$0.87 \:(0.01)$ & 
$0.38 \:(0.00)$ & 
$0.53 \:(0.00)$\\

Yes & Yes & No &
$0.60 \:(0.06)$ & 
$0.84 \:(0.08)$ & 
$0.69 \:(0.01)$\\

Yes & Yes & Yes & 
$0.88 \:(0.03)$ & 
$0.58 \:(0.01)$ & 
$0.70 \:(0.01)$\\

\bottomrule
\end{tabular}%

}
\captionof{table}{Ablation results on varying chain-of-thought (CoT.), summarization (Sum.), and reflection (Ref.) settings for Nepali. All results are an average over 5 runs with standard deviations shown in parentheses.}
\label{tab:prompt_ablations_nepali}

\vspace{0.5cm}
{
\begin{tabular}{cccccc}
\toprule
\multicolumn{3}{c}{\textbf{Setting}} &  \multicolumn{3}{c}{\textbf{Performance}} \\
\cmidrule(lr){1-3}
\cmidrule(lr){4-6}
\textbf{CoT.} & \textbf{Sum.} & \textbf{Ref.} & \textbf{P} & \textbf{R} & \textbf{F1} \\
\midrule

No & No & No & 
$0.39 \:(0.03)$ & 
$1.00 \:(0.00)$ & 
$0.56 \:(0.03)$\\

No & No & Yes & 
$1.00 \:(0.00)$ & 
$0.76 \:(0.00)$ & 
$0.86 \:(0.00)$\\

No & Yes & No & 
$0.37 \:(0.03)$ & 
$0.98 \:(0.02)$& 
$0.53 \:(0.03)$\\

No & Yes & Yes & 
$0.86 \:(0.03)$ & 
$0.72 \:(0.02)$ & 
$0.79 \:(0.02)$\\

Yes & No & No & 
$0.46 \:(0.04)$ & 
$0.99 \:(0.02)$ &
$0.62 \:(0.04)$\\

Yes & No & Yes & 
$1.00 \:(0.00)$ & 
$0.81 \:(0.00)$ & 
$0.89 \:(0.00)$\\

Yes & Yes & No &
$0.52 \:(0.04)$ & 
$0.94 \:(0.06)$ & 
$0.67 \:(0.03)$\\

Yes & Yes & Yes & 
$0.89 \:(0.02)$ & 
$0.71 \:(0.00)$ & 
$0.79 \:(0.01)$\\
\bottomrule
\end{tabular}%
}
\caption{Ablation results on varying chain-of-thought (CoT.), summarization (Sum.), and reflection (Ref.) settings for Spanish. All results are an average over 5 runs with standard deviations shown in parentheses.}
\label{tab:prompt_ablations_colombian}
\end{table}

NewsSerow does much better than multilingual encoder models when the latter are trained in a few-shot setting and that it achieves comparable performance to the multilingual models when they are trained with a much larger amount of data. To study this more closely, we consider a setting in which labeled English language data are easily available while the target-language data is much harder to acquire. This is a reasonable assumption especially for a lower-resourced language such as Nepali. We measure the performance of XLM-R on the Nepali language test set when finetuned on 1647 English examples and a variable number of Nepali examples. We benchmark this against NewsSerow's performance using 10 Nepali examples (Figure \ref{fig:number_ex_xlmr}). It can be observed that in addition to English examples, XLM-R requires at least 90 Nepali-language examples to reach the performance obtained by NewsSerow. Even then, NewsSerow achieves less variance compared to XLM-R.

\subsubsection{Where Does NewsSerow Make Mistakes?} One of the failure modes of NewsSerow is making a tenuous connection between an article and environmental conservation to output a false positive. For example, for an article about climbing preparation, it outputs: \textit{``The article talks about the preparation for climbing Annapurna and Mt. Everest, which are both related to environment conservation as they are conservation sites."} 
While mountains are conservation sites, the subject of the referred article is a climb. As another example, for an article about the architectural style for new construction in an ancient city, the model responds by saying \textit{``The article talks about the construction of houses in an ancient city of Panauti in a way that preserves the traditional style of architecture. This news is related to cultural and environmental conservation."} In both cases, the model correctly captures the meaning of the article but incorrectly labels it as being relevant. Often, though not always, the self-reflection step is able to correct this. Even in cases where it is not corrected, the response provides value to a domain expert by being interpretable and easy to spot as being incorrect.

\subsection{Ablation Studies} \label{subsec:ablations}
In this section, we first study the effects of varying the number of in-context examples and then discuss ablations on the prompt engineering features of NewsSerow: zero-shot summarization of article text, chain-of-thought reasoning in the classification prompt, and reflection of the classification response. The effects of these three features are summarized in Tables~\ref{tab:prompt_ablations_nepali} and~\ref{tab:prompt_ablations_colombian}, where we illustrate performance variations caused by switching each of these features on/off.

\subsubsection{Effect of the Number of In-Context Examples:}
As shown in Figure~\ref{fig:number_ice}, increasing the number of in-context examples from 2 to 10 gives a slight upward trend in performance across both languages. Additionally, while there is not a clean trend between the number of examples and variance, we observe that instances of high variance typically occur when a lower number of examples is used. A higher number of example stabilizes performance.

\subsubsection{Effect of Chain-of-Thought Reasoning:} Recall from Section \ref{subsec:classification} that we model chain-of-thought reasoning by giving explanations for true labels of the example articles. We observe (Tables~\ref{tab:prompt_ablations_nepali} and~\ref{tab:prompt_ablations_colombian}) that using chain-of-thought explanations while keeping the other settings constant always improves F1-score for both Nepali and Spanish. In the case of Nepali, this can cleanly be attributed to an increase in recall that outpaces a slight drop in precision. In the case of the Colombian dataset, recall on average stays similar while precision gets a slight boost, thus improving the F1 score. 

\subsubsection{Effect of the Summarization Module:} Recall that we use article summaries to efficiently utilize the limited context window of the classification prompt while ensuring that the salient points of the article are captured.

Using the first few sentences of a text is often used as a strong extractive summarization baseline \cite{narayan-2018-dont, nenkova-2005-automatic}. Consequently, to study the efficacy of the summarization module, we compare it to a baseline where we use the first 3 sentences of the article as a summary. In Tables~\ref{tab:prompt_ablations_nepali} and~\ref{tab:prompt_ablations_colombian}, the summarization column takes the value \texttt{yes} if the summarization module is used to generate a zero-shot summary and \texttt{no} if the first three sentences of the article are used as an extractive summary.

We observe that for Nepali, the summarization module always improves accuracy and F1 score, by boosting recall without causing a precision drop. For Spanish, recall generally decreases on using the zero-shot summary, whereas precision does not show a neat pattern. 

\subsubsection{Effect of Using Reflection:} 
Recall from Section \ref{subsec:reflection} that we use reflection only when the prediction made by the classification module is positive. As such, reflection increases precision at the cost of recall. In this, we find reflection to be an effective precision-recall switch that can be used by the non-profits to control for the precision-recall tradeoff based on their requirements and resource availability.

\section{Deployment}
NewsSerow has been deployed by WWF in Nepal and Colombia since April 2023. Every week, NewsSerow scrapes articles as described in Section~\ref{subsec:data}, runs the classification pipeline on the retrieved articles, and shares the lists of articles predicted to be relevant to conservation with WWF. We have used the weekly feedback from our collaborators from WWF to make iterative improvements to NewsSerow. This paper describes the latest version of NewsSerow, which was deployed in August 2023.    

We present the results on NewsSerow's performance on data labeled by WWF from April to July 2023 in Table~\ref{tab:deploy}. Because the objective of this earlier process was to iterate on previous versions of our pipeline, these evaluation points from April to July 2023 are not necessarily representative of the distribution of articles NewsSerow observes in a regular week. In particular, through our weekly pipeline, we have been requesting labels for only the lists of articles that were classified as being positive by (earlier iterations of) NewsSerow. And hence the precision, recall, and F1 score are calculated for this subset of the data only.\footnote{Note that the data described in Section \ref{subsec:data} for all the offline experiments are unbiased since they are made up of a random sample of articles scraped between December 2022 and March 2023.}  Labeling the whole dataset would be too burdensome for our partners at WWF and would go against the initial motivation of this project. Additionally, since we have used our observations from this weekly exchange of results in developing NewsSerow, it can be considered a pseudo-validation set. 
Thus, the evaluation metrics reported below are not exactly comparable with those from the offline experiments in Section~\ref{sec:experiments}.
Nonetheless, the results still provide valuable insights into the real-world performance of NewsSerow, as below.

As shown in Table~\ref{tab:deploy}, over the 8 weeks of deployment so far, NewsSerow achieves great performance across both languages. It achieves F1 scores of 0.64 and 0.67 on Nepali and Colombian articles respectively. The variations in results across weeks, especially for Nepali, are likely due to the limited number of data points per week for Nepali and is less of an issue with Spanish which has more articles weekly. Notably, the performance metrics are not as good as those in the offline experiments. As described previously, the offline and deployment datasets do not have the same distribution. Further, there is a time shift in the training and testing distributions for the deployment results but not for the offline experiments. Nevertheless, the F1 scores for both languages in deployment is still higher than the other few-shot models.


\begin{table}[t]
\small
\setlength{\tabcolsep}{4pt}  
\centering
{
\begin{tabular}{crcccrccc}
\toprule
 &  \multicolumn{4}{c}{\textbf{Nepal}} &  \multicolumn{4}{c}{\textbf{Spanish (Colombia)}}\\

\cmidrule(lr){2-5}
\cmidrule(lr){6-9}

\textbf{Week} & \textbf{\# Ex.} & \textbf{P} & \textbf{R} & \textbf{F1} 

& \textbf{\# Ex.} & \textbf{P} & \textbf{R} & \textbf{F1}
\\
\midrule

1 & 10 & 
$0.5 $ & 
$1.00$ & 
$0.67$ 

& 28 & 0.57 & 1.00 & 0.73\\

2 & 9 & 
$0.00$ & 
$0.00$ & 
$0.00$

&22 & 1.00 & 0.70 & 0.82\\

3 & 21 & 
$1.00$ & 
$0.33$& 
$0.50$

& 28 & 0.5 & 0.8 & 0.62\\

4 & 9 & 
$1.00$ & 
$0.75$ & 
$0.86$

& 26 & 0.63 & 0.71 & 0.67\\

5 & 9 &
$0.50$ & 
$0.50$ &
$0.50$

& 35 & 0.67 & 0.80 & 0.73\\

6 & 5 & 
$0.50$ & 
$1.00$ & 
$0.67$

& 23 & 0.38 & 0.50 & 0.43\\

7 & 8 &
$1.00$ & 
$1.00$ & 
$1.00$

& & & & \\

8 & 13 & 
$1.00$ & 
$0.38$ & 
$0.55$

& & & & \\
\midrule

\textbf{Aggr.} & $\mathbf{84}$ & 
$\mathbf{0.77}$ & 
$\mathbf{0.55}$ & 
$\mathbf{0.64}$ & 

$\mathbf{162}$ &
$\mathbf{0.61}$ & 
$\mathbf{0.73}$ & 
$\mathbf{0.67}$\\

\bottomrule
\end{tabular}%
}
\caption{Weekly results of the NewsSerow deployed model from April to July 2023. For Nepali, we have 84 data points collected over 8 weeks. For Spanish, we have 162 data points collected over 6 weeks.}
\label{tab:deploy}
\end{table}

\section{Conclusion} \label{sec:conclusion}
In this paper, we present NewsSerow, a multilingual few-shot framework that utilizes large language models (LLMs) for identifying news content relevant to conservation efforts. We illustrate that LLMs with well-engineered prompts can utilize a handful of illustrative examples to perform on par with or better than fine-tuned baselines that require training on much larger datasets. In particular, this holds true also for lower-resourced languages such as Nepali, which are typically not as well represented in the corpora used to pre-train LLMs as higher-resourced languages such as English. Consequently, the framework can easily be adapted by conservation-focused non-profits working in local languages anywhere in the world, and save significant amounts of human effort and time that goes into manual media monitoring.

\section*{Acknowledgements}
We thank Kanishka Bhambhani and Pratik Joshi for writing the scrapers to collect the Nepali news data used in this paper. The work is supported in part by NSF grant IIS-2046640 (CAREER) and a research grant from Google. Zheyuan Ryan Shi was supported by a CMU Presidential Fellowship and Siebel Scholarship and did this work while at CMU. Lei Li’s work is partially supported by a gift grant from Apple Inc.
\bibliography{aaai24}

\begin{thebibliography}{22}
\providecommand{\natexlab}[1]{#1}

\bibitem[{Brown et~al.(2020)Brown, Mann, Ryder, Subbiah, Kaplan, Dhariwal, Neelakantan, Shyam, Sastry, Askell, Agarwal, Herbert-Voss, Krueger, Henighan, Child, Ramesh, Ziegler, Wu, Winter, Hesse, Chen, Sigler, Litwin, Gray, Chess, Clark, Berner, McCandlish, Radford, Sutskever, and Amodei}]{brown-2020-language}
Brown, T.; Mann, B.; Ryder, N.; Subbiah, M.; Kaplan, J.~D.; Dhariwal, P.; Neelakantan, A.; Shyam, P.; Sastry, G.; Askell, A.; Agarwal, S.; Herbert-Voss, A.; Krueger, G.; Henighan, T.; Child, R.; Ramesh, A.; Ziegler, D.; Wu, J.; Winter, C.; Hesse, C.; Chen, M.; Sigler, E.; Litwin, M.; Gray, S.; Chess, B.; Clark, J.; Berner, C.; McCandlish, S.; Radford, A.; Sutskever, I.; and Amodei, D. 2020.
\newblock Language Models are Few-Shot Learners.
\newblock In Larochelle, H.; Ranzato, M.; Hadsell, R.; Balcan, M.; and Lin, H., eds., \emph{Advances in Neural Information Processing Systems}, volume~33, 1877--1901. Curran Associates, Inc.

\bibitem[{Conneau et~al.(2020)Conneau, Khandelwal, Goyal, Chaudhary, Wenzek, Guzm{\'a}n, Grave, Ott, Zettlemoyer, and Stoyanov}]{conneau-2020-unsupervised}
Conneau, A.; Khandelwal, K.; Goyal, N.; Chaudhary, V.; Wenzek, G.; Guzm{\'a}n, F.; Grave, E.; Ott, M.; Zettlemoyer, L.; and Stoyanov, V. 2020.
\newblock Unsupervised Cross-lingual Representation Learning at Scale.
\newblock In \emph{Proceedings of the 58th Annual Meeting of the Association for Computational Linguistics}, 8440--8451. Online: Association for Computational Linguistics.

\bibitem[{Devlin et~al.(2019)Devlin, Chang, Lee, and Toutanova}]{devlin-2019-bert}
Devlin, J.; Chang, M.-W.; Lee, K.; and Toutanova, K. 2019.
\newblock {BERT}: Pre-training of Deep Bidirectional Transformers for Language Understanding.
\newblock In \emph{Proceedings of the 2019 Conference of the North {A}merican Chapter of the Association for Computational Linguistics: Human Language Technologies, Volume 1 (Long and Short Papers)}, 4171--4186. Minneapolis, Minnesota: Association for Computational Linguistics.

\bibitem[{Gilardi, Alizadeh, and Kubli(2023)}]{Gilardi2023ChatGPTOC}
Gilardi, F.; Alizadeh, M.; and Kubli, M. 2023.
\newblock ChatGPT outperforms crowd workers for text-annotation tasks.
\newblock \emph{Proceedings of the National Academy of Sciences of the United States of America}, 120.

\bibitem[{Hosseini and Coll~Ardanuy(2020)}]{hosseini-coll-ardanuy}
Hosseini, K.; and Coll~Ardanuy, M. 2020.
\newblock Data Study Group Final Report: WWF.

\bibitem[{Jain et~al.(2023)Jain, Keshava, Mysore~Sathyendra, Fernandes, Liu, Neubig, and Zhou}]{jain-2023-multi}
Jain, S.; Keshava, V.; Mysore~Sathyendra, S.; Fernandes, P.; Liu, P.; Neubig, G.; and Zhou, C. 2023.
\newblock Multi-Dimensional Evaluation of Text Summarization with In-Context Learning.
\newblock In \emph{Findings of the Association for Computational Linguistics: ACL 2023}. Toronto, Canada: Association for Computational Linguistics.

\bibitem[{Jang(2023)}]{jang2023reflection}
Jang, E. 2023.
\newblock Can LLMs Critique and Iterate on Their Own Outputs?
\newblock \emph{evjang.com}.

\bibitem[{Kadavath et~al.(2022)Kadavath, Conerly, Askell, Henighan, Drain, Perez, Schiefer, Hatfield-Dodds, DasSarma, Tran-Johnson, Johnston, El-Showk, Jones, Elhage, Hume, Chen, Bai, Bowman, Fort, Ganguli, Hernandez, Jacobson, Kernion, Kravec, Lovitt, Ndousse, Olsson, Ringer, Amodei, Brown, Clark, Joseph, Mann, McCandlish, Olah, and Kaplan}]{kadavath-2022-language}
Kadavath, S.; Conerly, T.; Askell, A.; Henighan, T.; Drain, D.; Perez, E.; Schiefer, N.; Hatfield-Dodds, Z.; DasSarma, N.; Tran-Johnson, E.; Johnston, S.; El-Showk, S.; Jones, A.; Elhage, N.; Hume, T.; Chen, A.; Bai, Y.; Bowman, S.; Fort, S.; Ganguli, D.; Hernandez, D.; Jacobson, J.; Kernion, J.; Kravec, S.; Lovitt, L.; Ndousse, K.; Olsson, C.; Ringer, S.; Amodei, D.; Brown, T.; Clark, J.; Joseph, N.; Mann, B.; McCandlish, S.; Olah, C.; and Kaplan, J. 2022.
\newblock Language Models (Mostly) Know What They Know.
\newblock arXiv:2207.05221.

\bibitem[{Keh et~al.(2023)Keh, Shi, Patterson, Bhagabati, Dewan, Gopala, Izquierdo, Mallick, Sharma, Shrestha, and Fang}]{keh-2023-newspanda}
Keh, S.~S.; Shi, Z.~R.; Patterson, D.~J.; Bhagabati, N.; Dewan, K.; Gopala, A.; Izquierdo, P.; Mallick, D.; Sharma, A.; Shrestha, P.; and Fang, F. 2023.
\newblock NewsPanda: Media Monitoring for Timely Conservation Action.
\newblock \emph{Proceedings of the AAAI Conference on Artificial Intelligence}, 37(13): 15528--15536.

\bibitem[{Kojima et~al.(2022)Kojima, Gu, Reid, Matsuo, and Iwasawa}]{Kojima2022LargeLM}
Kojima, T.; Gu, S.~S.; Reid, M.; Matsuo, Y.; and Iwasawa, Y. 2022.
\newblock Large Language Models are Zero-Shot Reasoners.
\newblock \emph{ArXiv}, abs/2205.11916.

\bibitem[{Li et~al.(2023)Li, Zhang, Dubois, Taori, Gulrajani, Guestrin, Liang, and Hashimoto}]{alpaca_eval}
Li, X.; Zhang, T.; Dubois, Y.; Taori, R.; Gulrajani, I.; Guestrin, C.; Liang, P.; and Hashimoto, T.~B. 2023.
\newblock AlpacaEval: An Automatic Evaluator of Instruction-following Models.
\newblock \url{https://github.com/tatsu-lab/alpaca_eval}.

\bibitem[{Liu et~al.(2021)Liu, Yuan, Fu, Jiang, Hayashi, and Neubig}]{DBLP:journals/corr/abs-2107-13586}
Liu, P.; Yuan, W.; Fu, J.; Jiang, Z.; Hayashi, H.; and Neubig, G. 2021.
\newblock Pre-train, Prompt, and Predict: {A} Systematic Survey of Prompting Methods in Natural Language Processing.
\newblock \emph{CoRR}, abs/2107.13586.

\bibitem[{Liu et~al.(2019)Liu, Ott, Goyal, Du, Joshi, Chen, Levy, Lewis, Zettlemoyer, and Stoyanov}]{liu-2019-roberta}
Liu, Y.; Ott, M.; Goyal, N.; Du, J.; Joshi, M.; Chen, D.; Levy, O.; Lewis, M.; Zettlemoyer, L.; and Stoyanov, V. 2019.
\newblock RoBERTa: A Robustly Optimized BERT Pretraining Approach.
\newblock arXiv:1907.11692.

\bibitem[{Loshchilov and Hutter(2019)}]{loshchilov-2018-decoupled}
Loshchilov, I.; and Hutter, F. 2019.
\newblock Decoupled Weight Decay Regularization.
\newblock In \emph{International Conference on Learning Representations}.

\bibitem[{Narayan, Cohen, and Lapata(2018)}]{narayan-2018-dont}
Narayan, S.; Cohen, S.~B.; and Lapata, M. 2018.
\newblock Don{'}t Give Me the Details, Just the Summary! Topic-Aware Convolutional Neural Networks for Extreme Summarization.
\newblock In \emph{Proceedings of the 2018 Conference on Empirical Methods in Natural Language Processing}, 1797--1807. Brussels, Belgium: Association for Computational Linguistics.

\bibitem[{Nenkova(2005)}]{nenkova-2005-automatic}
Nenkova, A. 2005.
\newblock Automatic Text Summarization of Newswire: Lessons Learned from the Document Understanding Conference.
\newblock In \emph{AAAI Conference on Artificial Intelligence}.

\bibitem[{Nye et~al.(2021)Nye, Andreassen, Gur{-}Ari, Michalewski, Austin, Bieber, Dohan, Lewkowycz, Bosma, Luan, Sutton, and Odena}]{scratchpad}
Nye, M.~I.; Andreassen, A.~J.; Gur{-}Ari, G.; Michalewski, H.; Austin, J.; Bieber, D.; Dohan, D.; Lewkowycz, A.; Bosma, M.; Luan, D.; Sutton, C.; and Odena, A. 2021.
\newblock Show Your Work: Scratchpads for Intermediate Computation with Language Models.
\newblock \emph{CoRR}, abs/2112.00114.

\bibitem[{OpenAI(2023)}]{OpenAI2023GPT4TR}
OpenAI. 2023.
\newblock GPT-4 Technical Report.
\newblock \emph{ArXiv}, abs/2303.08774.

\bibitem[{Shinn et~al.(2023)Shinn, Cassano, Labash, Gopinath, Narasimhan, and Yao}]{shinn-2023-reflexion}
Shinn, N.; Cassano, F.; Labash, B.; Gopinath, A.; Narasimhan, K.; and Yao, S. 2023.
\newblock Reflexion: Language Agents with Verbal Reinforcement Learning.
\newblock arXiv:2303.11366.

\bibitem[{Touvron et~al.(2023)Touvron, Martin, Stone, Albert, Almahairi, Babaei, Bashlykov, Batra, Bhargava, Bhosale, Bikel, Blecher, Ferrer, Chen, Cucurull, Esiobu, Fernandes, Fu, Fu, Fuller, Gao, Goswami, Goyal, Hartshorn, Hosseini, Hou, Inan, Kardas, Kerkez, Khabsa, Kloumann, Korenev, Koura, Lachaux, Lavril, Lee, Liskovich, Lu, Mao, Martinet, Mihaylov, Mishra, Molybog, Nie, Poulton, Reizenstein, Rungta, Saladi, Schelten, Silva, Smith, Subramanian, Tan, Tang, Taylor, Williams, Kuan, Xu, Yan, Zarov, Zhang, Fan, Kambadur, Narang, Rodriguez, Stojnic, Edunov, and Scialom}]{Touvron2023Llama2O}
Touvron, H.; Martin, L.; Stone, K.~R.; Albert, P.; Almahairi, A.; Babaei, Y.; Bashlykov, N.; Batra, S.; Bhargava, P.; Bhosale, S.; Bikel, D.~M.; Blecher, L.; Ferrer, C.~C.; Chen, M.; Cucurull, G.; Esiobu, D.; Fernandes, J.; Fu, J.; Fu, W.; Fuller, B.; Gao, C.; Goswami, V.; Goyal, N.; Hartshorn, A.~S.; Hosseini, S.; Hou, R.; Inan, H.; Kardas, M.; Kerkez, V.; Khabsa, M.; Kloumann, I.~M.; Korenev, A.~V.; Koura, P.~S.; Lachaux, M.-A.; Lavril, T.; Lee, J.; Liskovich, D.; Lu, Y.; Mao, Y.; Martinet, X.; Mihaylov, T.; Mishra, P.; Molybog, I.; Nie, Y.; Poulton, A.; Reizenstein, J.; Rungta, R.; Saladi, K.; Schelten, A.; Silva, R.; Smith, E.~M.; Subramanian, R.; Tan, X.; Tang, B.; Taylor, R.; Williams, A.; Kuan, J.~X.; Xu, P.; Yan, Z.; Zarov, I.; Zhang, Y.; Fan, A.; Kambadur, M.; Narang, S.; Rodriguez, A.; Stojnic, R.; Edunov, S.; and Scialom, T. 2023.
\newblock Llama 2: Open Foundation and Fine-Tuned Chat Models.
\newblock \emph{ArXiv}, abs/2307.09288.

\bibitem[{Wei et~al.(2022)Wei, Wang, Schuurmans, Bosma, brian ichter, Xia, Chi, Le, and Zhou}]{wei-2022-chain}
Wei, J.; Wang, X.; Schuurmans, D.; Bosma, M.; brian ichter; Xia, F.; Chi, E.~H.; Le, Q.~V.; and Zhou, D. 2022.
\newblock Chain of Thought Prompting Elicits Reasoning in Large Language Models.
\newblock In Oh, A.~H.; Agarwal, A.; Belgrave, D.; and Cho, K., eds., \emph{Advances in Neural Information Processing Systems}.

\bibitem[{Welleck et~al.(2022)Welleck, Lu, West, Brahman, Shen, Khashabi, and Choi}]{Welleck2022GeneratingSB}
Welleck, S.; Lu, X.; West, P.; Brahman, F.; Shen, T.; Khashabi, D.; and Choi, Y. 2022.
\newblock Generating Sequences by Learning to Self-Correct.
\newblock \emph{ArXiv}, abs/2211.00053.

\end{thebibliography}


\end{document}